\documentclass{article}
\usepackage[preprint]{neurips_2024}

\usepackage[utf8]{inputenc}
\usepackage[T1]{fontenc}
\usepackage{hyperref}
\usepackage{url}
\usepackage{booktabs}
\usepackage{amsfonts}
\usepackage{amsmath}
\usepackage{nicefrac}
\usepackage{microtype}
\usepackage{graphicx}
\usepackage{xcolor}
\usepackage{subcaption}
\usepackage{multirow}
\usepackage{wrapfig}
\usepackage{float}

\graphicspath{{figures/}}

\title{Directional Routing in Transformers}

\author{
  Kevin Taylor \\
  \texttt{kevin.taylor1924@gmail.com}
}

\begin{document}

\maketitle

\begin{abstract}
We introduce \textbf{directional routing}, a lightweight mechanism that gives each transformer attention head learned suppression directions controlled by a shared router, at 3.9\% parameter cost. We train a 433M-parameter model alongside an identical baseline in a single run, then trace the resulting circuits through mechanistic interpretability.

Routing becomes the model's dominant computational pathway. Disabling it collapses factual recall to near-zero probability across all 8 test prompts and drops induction accuracy from 93.4\% to 0.0\%. Knocking out individual attention heads has negligible effect: the primary mover head's removal actually \textit{increases} target probability, and induction heads retain 98.6\% accuracy without their strongest member. The coordination mechanism is irreplaceable; the components it coordinates are not. The model also self-organizes, without explicit pressure, into two regimes: domain-adaptive routing in early layers and fixed syntactic pruning in late layers, where the least-varying layer is the most critical ($+42.6$ PPL when disabled). Routing reduces perplexity 31--56\% relative to the baseline, though downstream multiple-choice benchmarks do not yet reflect these gains.
\end{abstract}

\section{Introduction}
\label{sec:intro}

Transformers learn powerful representations but offer no built-in account of what those representations encode. Understanding them requires post-hoc tools: sparse autoencoders \citep{cunningham2024sparse, bricken2023monosemanticity}, probing classifiers \citep{belinkov2022probing}, and causal tracing \citep{meng2022locating}, each adding computational cost and producing explanations that approximate, rather than expose, the model's actual mechanism. Mixture-of-experts architectures \citep{shazeer2017outrageously, fedus2022switch, lepikhin2021gshard, jiang2024mixtral, dai2024deepseekmoe} provide some structural transparency through expert specialization, but at the price of large parameter overhead and complex routing logistics.

We propose \textbf{directional routing}: each attention head learns $K{=}4$ unit-norm direction vectors in head-space, and a shared 4-layer MLP router produces per-input weights that control how much of each directional component is removed from the head's output. The entire mechanism adds 3.9\% parameters (16.2M) and 0.02\% FLOPs to a standard transformer. No auxiliary routing loss is used; the router learns purely from the language modeling objective.

This paper asks: \textit{what does the model learn to do with routing?} The primary finding is that routing becomes the load-bearing computational mechanism. On both factual recall and induction, disabling routing collapses the circuit completely, while knocking out individual heads has negligible effect. The coordination mechanism matters; the components it coordinates do not.

\section{Method}
\label{sec:method}

\subsection{Architecture}

\begin{figure}[t]
    \centering
    \includegraphics[width=\textwidth]{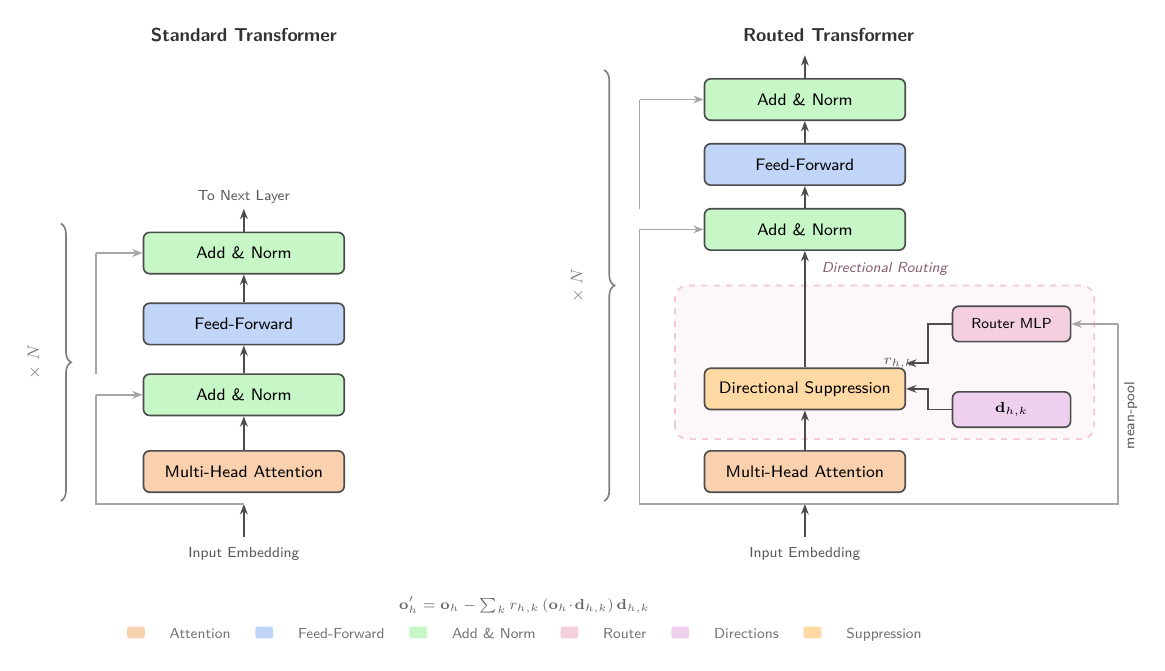}
    \caption{Standard transformer block (left) vs.\ routed transformer block (right). The routing mechanism (dashed box) adds three components: a shared router MLP that produces per-input weights $r_{h,k} \in [0,1]$ from the mean-pooled sequence, learned direction vectors $\mathbf{d}_{h,k}$, and a directional suppression step that removes selected components from each head's output. All other components are identical.}
    \label{fig:architecture}
\end{figure}

We augment a standard transformer \citep{vaswani2017attention} with three additions to the attention mechanism (Figure~\ref{fig:architecture}). The base model has $L{=}12$ layers, $H{=}12$ heads per layer, $d_{\text{model}}{=}1536$, and $d_{\text{head}}{=}128$.

\paragraph{Direction vectors.} Each head learns $K{=}4$ direction vectors $\mathbf{d}_{h,k} \in \mathbb{R}^{d_h}$, normalized to unit length during the forward pass. This adds $L \times H \times K \times d_h = 73{,}728$ parameters, a negligible fraction of the model.

\paragraph{Router.} A 4-layer MLP, shared across all heads within each layer, produces routing weights from a mean-pooled sequence representation:
\begin{equation}
    \mathbf{r} = \sigma\!\left(T \cdot \mathrm{MLP}_4\!\left(\frac{1}{n}\sum_{i=1}^{n} \mathbf{x}_i\right)\right) \in [0,1]^{H \times K}
    \label{eq:router}
\end{equation}
where $\mathbf{x}_i$ are residual stream vectors, $T{=}5.0$ is a temperature that pushes weights toward binary decisions, and $\sigma$ is the sigmoid function. Mean-pooling yields a single routing decision per sequence. The 12 router MLPs total 16.1M parameters (3.7\% of the model).

\paragraph{Directional suppression.} After scaled dot-product attention computes head output $\mathbf{o}_h$, we apply:
\begin{equation}
    \mathbf{o}'_h = \mathbf{o}_h - \sum_{k=1}^{K} r_{h,k} \cdot (\mathbf{o}_h \cdot \mathbf{d}_{h,k}) \, \mathbf{d}_{h,k}
    \label{eq:suppression}
\end{equation}
When $r_{h,k}{=}0$, no suppression occurs. When $r_{h,k}{=}1$, the component along $\mathbf{d}_{h,k}$ is fully removed. The router decides the suppression pattern for each input.

\subsection{Training}

Both models are trained identically: AdamW \citep{loshchilov2019decoupled} with cosine learning rate schedule, 20K steps on FineWeb \citep{penedo2024fineweb} (${\sim}2.5$B tokens), GPT-2 tokenizer \citep{radford2019language}. The routed model has 433M parameters; the baseline has 417M (3.9\% overhead). No auxiliary routing loss, load-balancing objective, or special initialization is used. The router learns what to suppress purely from the next-token prediction loss. We also trained a 26M-parameter variant of both models (8 layers, 8 heads, $K{=}8$) on TinyStories and FineWeb for ablation purposes; these results appear in \S\ref{sec:discussion}. All results are from a single training run per model; we do not report variance across seeds.

\section{Circuit Analysis}
\label{sec:circuits}

We analyze routing's role in two well-studied circuits, factual recall \citep{meng2022locating, geva2023dissecting} and induction \citep{olsson2022context}, and find the same pattern in both: routing is the single non-redundant component, while individual attention heads are interchangeable. This is an empirical observation about what the trained model learned to depend on. A model trained without routing would develop different circuits entirely, and we have not tested other circuit types.

\subsection{Factual Recall}

We probe with ``The capital of France is'' and measure the logit assigned to Paris. This is not a cherry-picked success case: Paris is not even the model's top prediction (``the'' is), and receives only $P = 0.00119$ under normal routing. What matters is the contrast. Setting all routing weights to zero collapses Logit(Paris) from $+4.73$ to $-6.21$, a swing of $10.94$ (Table~\ref{tab:routing-necessity}). The same collapse occurs on all 8 factual prompts tested (Japan $\to$ Tokyo, Germany $\to$ Berlin, gold $\to$ Au, and five others).

\begin{table}[t]
\centering
\caption{Routing's effect on factual logits (``The capital of France is $\to$ Paris''). Paris is not top-1 in any condition; routing controls whether it receives a positive logit at all.}
\label{tab:routing-necessity}
\begin{tabular}{lccc}
\toprule
Condition & P(Paris) & Logit(Paris) & Top Prediction \\
\midrule
Normal (learned routing) & 0.00119 & $+4.73$ & ``the'' \\
Routing OFF ($w{=}0$) & 0.00000 & $-6.21$ & ``put'' \\
Routing NEUTRAL ($w{=}0.5$) & 0.00084 & $+4.10$ & ``once'' \\
Routing FULL ($w{=}1$) & 0.00006 & $+5.34$ & ``unique'' \\
\bottomrule
\end{tabular}
\end{table}

Logit attribution via direct subtraction explains the mechanism (Table~\ref{tab:logit-main}). For each component class, we compute Logit(Paris) with the component active minus Logit(Paris) with it zeroed out. Raw attention heads \textit{oppose} the correct answer ($-4.94$ logits); routing contributes $+25.57$, concentrated in L8H2 ($+4.14$), L9H4 ($+3.17$), and L10H8 ($+2.55$). Because LayerNorm rescaling is nonlinear, these marginal contributions do not sum to the final logit ($+16.70$ vs.\ $+4.73$); the residual reflects LayerNorm compression and cross-term interactions. We interpret these as relative magnitudes rather than an exact additive decomposition. Routing's marginal effect is $5\times$ larger than the raw head effect and opposite in sign.

\begin{table}[t]
\centering
\caption{Marginal logit attribution for ``The capital of France is $\to$ Paris,'' computed by direct subtraction. Non-additive due to LayerNorm; interpret as relative magnitudes.}
\label{tab:logit-main}
\begin{tabular}{lc}
\toprule
Component & Marginal $\Delta$ Logit(Paris) \\
\midrule
Token embedding & $-0.12$ \\
All attention heads (raw, pre-routing) & $-4.94$ \\
Routing delta (suppression effect) & $+25.57$ \\
All MLPs & $-3.81$ \\
\bottomrule
\end{tabular}
\end{table}

Individual heads tell the same story. Knocking out the primary mover head for this circuit (L10H11) \textit{increases} $P(\text{Paris})$ to $1.22\times$ normal. Across all 8 prompts, mover head knockout averages $1.07\times$ normal while routing OFF averages $0.0000\times$. Head redundancy is well-established in standard transformers \citep{voita2019analyzing, michel2019sixteen, clark2019what}, but here the redundancy is total: the model has learned distributed pathways where no single head is necessary, while the routing mechanism that coordinates them is critical.

A natural objection is that a model trained with routing will depend on routing. This is true, but the model had the freedom to learn small, dispensable routing weights. Instead, it converged on total dependence: routing contributing $+25.57$ logits while raw heads contribute $-4.94$. The architecture incentivizes routing as the primary computational channel.

\subsection{Induction}

The pattern repeats on induction. Accuracy drops from 93.4\% to 0.0\% when routing is disabled, but knocking out the three identified induction heads (L6H1, L6H7, L3H4) retains 92.5\%, or 98.6\% of normal (Table~\ref{tab:induction}). Individual heads are interchangeable; routing is not.

\begin{table}[t]
\centering
\caption{Induction circuit analysis. Routing is essential; individual induction heads are not.}
\label{tab:induction}
\begin{tabular}{lc}
\toprule
Condition & Induction Accuracy \\
\midrule
Normal & 93.4\% \\
Routing OFF & 0.0\% \\
KO induction heads (L6H1, L6H7, L3H4) & 92.5\% \\
KO control heads & 86.6\% \\
\bottomrule
\end{tabular}
\end{table}

\section{Emergent Two-Regime Architecture}
\label{sec:two-regime}

The circuit analysis establishes that routing is load-bearing. But do all layers use routing the same way, or do different layers develop different strategies? Without any architectural pressure or auxiliary loss, the model self-organizes into two distinct operating regimes.

\subsection{Domain-Adaptive Early Layers}

Layer~0 contains 10/12 specialist heads (83\%), where specialists are heads in the top quartile of cross-domain routing variance. The routing variance at Layer~0 is 0.070, the highest of any layer by a wide margin. PCA of router hidden states at Layer~0 shows clean separation of math, code, prose, and factual domains, and SVD of the router's first-layer weights reveals sensitivity to semantic content words (``math,'' ``programs,'' ``behavior''). Layer~3 is the secondary specialist layer (9/12 heads). Per-head routing statistics are in Appendix~\ref{app:head-stats}.

\subsection{Syntactic Pruning in Late Layers}

Layers 7--9 present the opposite pattern: zero specialist heads. Routing variance at Layer~9 is 0.00055, $127\times$ lower than Layer~0. Direction vectors in these layers target syntactic features (punctuation, articles, conjunctions, discourse markers), and Layer~9's routing weights never exceed 0.9 for any direction. It never fully suppresses anything.

Yet Layer~9 is the most critical layer in the model. Disabling its routing causes $+42.6$ PPL (Table~\ref{tab:knockout}), more than double the next-worst layer (L8 at $+15.3$). Layers~0--1 actually \textit{improve} slightly when routing is disabled ($-0.9$ and $-1.1$ PPL), suggesting the early specialist routing mildly overrides useful head outputs.

\begin{table}[t]
\centering
\caption{Per-layer routing knockout. Disabling routing one layer at a time. Baseline dynamic routing: PPL 15.0.}
\label{tab:knockout}
\begin{tabular}{lcclcc}
\toprule
Layer & PPL & $\Delta$ PPL & Layer & PPL & $\Delta$ PPL \\
\midrule
0  & 14.1 & $-0.9$ & 6  & 21.2 & $+6.2$ \\
1  & 13.9 & $-1.1$ & 7  & 16.5 & $+1.5$ \\
2  & 15.2 & $+0.2$ & 8  & 30.3 & $+15.3$ \\
3  & 17.4 & $+2.4$ & 9  & 57.6 & $+42.6$ \\
4  & 18.1 & $+3.1$ & 10 & 18.1 & $+3.1$ \\
5  & 17.5 & $+2.5$ & 11 & 17.5 & $+2.5$ \\
\bottomrule
\end{tabular}
\end{table}

Cross-domain routing swaps confirm the split. Swapping routing weights between code and prose at early layers increases loss by $+5.5$; swapping at late layers increases loss by $<0.2$. Late-layer routing is domain-agnostic, nearly interchangeable across input types. The layers that vary most across domains (L0, L3) are relatively dispensable; the layer that varies least (L9) is the most critical. Domain adaptation is useful but not load-bearing. The fixed syntactic pruning in late layers is what the model cannot do without.

\section{Efficiency}
\label{sec:efficiency}

We now turn to aggregate performance. Routing reduces perplexity substantially but does not improve multiple-choice benchmarks.

\subsection{Domain Perplexity}

\begin{table}[t]
\centering
\caption{Domain perplexity on hand-written evaluation passages (3 passages per domain; Code: 570 tokens, Math: 252, Prose: 253, Factual: 189; 1{,}264 total). All models evaluated on identical text. Pythia-410M \citep{biderman2023pythia}: 405M params, 300B tokens from The Pile, NeoX tokenizer. Our models: 2.5B tokens from FineWeb, GPT-2 tokenizer ($120\times$ less data). PPL is tokenizer-sensitive; cross-model comparisons are directional only.}
\label{tab:perplexity}
\begin{tabular}{lcccc}
\toprule
Domain & Pythia-410M & Baseline (417M) & Routed (433M) & $\Delta$ \\
\midrule
Code & 2.4 & 26.3 & 18.1 & $-31.2$\% \\
Math & 16.2 & 66.5 & 29.2 & $-56.1$\% \\
Prose & 17.2 & 57.3 & 25.0 & $-56.4$\% \\
Factual & 9.9 & 20.6 & 10.1 & $-51.0$\% \\
\midrule
Overall & 11.4 & 42.7 & 20.6 & $-51.8$\% \\
\bottomrule
\end{tabular}
\end{table}

Routing reduces perplexity by 31--56\% relative to the baseline across all four domains (Table~\ref{tab:perplexity}). We include Pythia-410M as a directional reference, but stress that this comparison is confounded: the models use different tokenizers (GPT-2 vs.\ NeoX), different training data (FineWeb vs.\ The Pile), and perplexity is tokenizer-sensitive. Our baseline is weak, trained on $120\times$ less data, so the 31--56\% reduction reflects routing's marginal value on a data-starved model. Whether similar gains hold under data-matched conditions is an open question requiring a controlled comparison (same tokens, 3.9\% more parameters allocated to FFN or extra heads instead of routing).

\subsection{Overhead}

The routing mechanism adds 16.2M parameters (3.9\%), of which 16.1M are in the router MLPs and 74K in direction vectors. FLOPs overhead is 0.02\%, since the router MLP is small relative to attention and feed-forward computation. Wall-clock throughput at sequence length 1024 drops from 3,651 to 3,210 tokens per second (13.7\% overhead) due to a sequential dependency: the mean-pooled representation must be computed before routing weights are available, preventing full pipeline overlap. At shorter sequences (128 tokens), overhead rises to 109\% because the router's fixed cost dominates. This is an implementation limitation; the router computation could be overlapped with early attention layers in a fused kernel.

\subsection{Benchmarks}

Perplexity gains do not transfer to multiple-choice benchmarks. Across 7 benchmarks (500 examples each; HellaSwag, ARC-Easy/Challenge, PIQA, WinoGrande, BoolQ, LAMBADA), the routed model wins 1 and loses 6. Six of seven deltas fall within approximate 95\% CIs ($\pm 6\%$); BoolQ ($-7.0\%$) is borderline significant. Full results with confidence intervals are in Appendix~\ref{app:ci}. At sub-billion parameter scales, perplexity and multiple-choice accuracy appear to measure different capabilities.

\section{Interpretability}
\label{sec:interpretability}

The 576 learned direction vectors ($12 \times 12 \times 4$) are interpretable as a byproduct of training, requiring no post-hoc decomposition. They are complementary to sparse autoencoders, not competitive: SAEs find 10K--100K features across the full residual stream, while our 576 directions exist only in attention outputs and cover a narrower, coarser slice of the model's computation. The advantage is zero additional cost; the limitation is scope.

\subsection{Vocabulary Projection}

Across all 576 directions, automatic categorization (matching top-10 vocabulary tokens against keyword lists) yields: 88.2\% content tokens, 4.3\% articles, 2.4\% punctuation, 2.1\% conjunctions, 1.0\% numbers, 1.0\% prepositions, 0.9\% pronouns. The labels are human-readable: L8H7K2 maps to ``and, period, comma, but'' (conjunctions and punctuation), L11H7K2 maps to ``period, close-paren-period, close-bracket-period'' (sentence boundaries), L11H5K1 maps to ``For, Also, also'' (discourse transitions). The syntactically labeled minority is concentrated in late layers, consistent with the two-regime structure of \S\ref{sec:two-regime}.

To obtain these labels, we embed each direction $\mathbf{d}_{h,k} \in \mathbb{R}^{d_h}$ in the full model dimension (zero-padding outside the head's slice), project through the attention output matrix $W_{\text{out}}$, and compute dot products with every row of the language model head: $\text{scores} = (W_{\text{out}} \, \tilde{\mathbf{d}}_{h,k}) \cdot W_{\text{LM}}^\top \in \mathbb{R}^{|\mathcal{V}|}$. The top-10 tokens by score label each direction. These labels are causally meaningful: overriding the routing weight for the 25 article-encoding directions shifts $P(\text{article as next token})$ by $-6.2\%$ on held-out text ($n{=}50$ sequences), confirming that routing weights modulate token-category probabilities through a single scalar per direction.

\subsection{Domain Classification}

The routing weight vector for a given input acts as a domain fingerprint. As a qualitative demonstration, nearest-centroid classification on a small sample ($n{=}24$, 6 per domain) achieves 23/24 correct on 4-domain classification (math, code, prose, factual). This sample is too small for statistical claims (the 95\% CI spans roughly $[0.79, 1.0]$), but PCA of routing vectors shows clean domain clusters, with code maximally distant from other domains (L2 = 0.479 to prose vs.\ 0.303 between math and factual). The more rigorous evidence for domain-dependent routing is the quantified variance analysis in \S\ref{sec:two-regime}.

\section{Ablation Studies}
\label{sec:ablations}

Three ablations characterize the nature of routing's contribution. First, at inference time, replacing the learned router with a fixed weight $w$ applied uniformly to all directions produces PPL 3,575 at the optimal $w{=}0.6$ (found by grid search), versus PPL 15.0 with dynamic routing, a $238\times$ gap. This is an inference-time intervention on the trained model, not a comparison to a model trained from scratch with fixed routing. The gap shows the trained model's circuits depend on input-specific decisions; it does not show that fixed routing is inherently inferior.

Second, the effects of routing are roughly additive across layers. Setting layers 7--8--9 to neutral routing ($w{=}0.5$) increases loss by $+0.067$; disabling all 12 layers causes $+3.87$. There are no strong synergistic interactions between routing at different layers.

Third, CKA similarity \citep{kornblith2019similarity} between the routed and baseline models is 0.95--0.99 at every layer. Routing does not change what the model learns. It changes how those representations are processed, sharpening predictions (top-1 probability: 0.42 vs.\ 0.31) by selectively removing predictable syntactic components from attention outputs.

\section{Discussion}
\label{sec:discussion}

Language models trained on diverse data face a fundamental tension. A model that sees math, code, prose, and factual text must pack all of these capabilities into a shared set of parameters. Prior work on superposition \citep{elhage2022superposition} has shown that models represent more features than they have dimensions for, encoding them in overlapping directions that interfere with each other. When a model processes a math problem, the neural pathways it uses for mathematical reasoning carry residual activation from features learned for prose, code, and everything else. This cross-domain interference is noise.

Directional routing can be understood as a mechanism for managing this interference. Rather than adding new parameters to represent new features, it gives the model a way to selectively suppress the features that are irrelevant to the current input. When the router detects a math problem, it suppresses prose-related and code-related components in the attention outputs; when it detects code, it suppresses the rest. The 31--56\% perplexity reduction is, under this view, the model denoising itself: removing the cross-domain interference that an unrouted model has no choice but to tolerate. The two-regime architecture (\S\ref{sec:two-regime}) supports this interpretation. Early layers, where domain identity is most useful for deciding what to suppress, show the highest routing variance across domains. Late layers, where the remaining noise is syntactic rather than semantic, apply a near-constant pruning pattern to all inputs. The model has learned, without supervision, that different kinds of noise require different suppression strategies.

The 31--56\% perplexity reduction does not transfer to multiple-choice benchmarks, and the output distribution data explains why. Routing sharpens the model's predictions without changing its representations: CKA similarity between the routed and baseline models is 0.95--0.99 at every layer, yet the routed model concentrates more probability on its top predictions (top-1 probability 0.42 vs.\ 0.31, output entropy 3.31 vs.\ 4.23). Routing is a better \textit{decoder} of the same underlying knowledge, not a source of new knowledge. Perplexity rewards confidence on tokens the model already partially knows; multiple-choice benchmarks test whether the model knows the answer at all. The two metrics measure different things, and routing helps only the first.

The vocabulary projection (\S\ref{sec:interpretability}) and logit attribution (\S\ref{sec:circuits}) together suggest a coherent picture of what routing does. Late-layer directions target predictable, low-information features: punctuation, articles, conjunctions, discourse markers. On factual recall, raw attention heads \textit{oppose} the correct answer ($-4.94$ logits to Paris), while routing overrides this opposition ($+25.57$). Attention heads carry both signal and noise; routing strips the noise. The result is not new information but a cleaner signal. This is consistent with the CKA finding: routing does not change what the model represents, only how it reads those representations out.

Standard circuit analysis in mechanistic interpretability focuses on identifying important individual components: induction heads \citep{olsson2022context}, mover heads, name mover heads \citep{wang2023interpretability}. Our results complicate this picture. In the routed model, no individual head is necessary for either factual recall or induction ($1.07\times$ normal on knockout). What matters is the coordination layer that modulates all heads simultaneously. This does not invalidate head-level analysis in standard transformers, but it suggests that architectures with explicit coordination mechanisms shift computational importance from individual components to the coordinator. Whether standard transformers exhibit a similar pattern through implicit coordination (e.g., via the residual stream) is an open question.

In a separate 6K-step training run, the routed model reached the baseline's final perplexity (PPL 16.3) by step 4,500, a $1.3\times$ convergence speedup. By step 6,000, the routed model achieved PPL 12.6, 22.4\% lower than the baseline at the same step count. This suggests routing provides an inductive bias that accelerates early training, though we cannot determine from one run whether this advantage persists or saturates at longer training horizons.

Routing benefit decays monotonically with token position: $+1.35$ log-prob at positions 0--9, declining to $+0.19$ at positions 100+. This follows directly from the mean-pooling bottleneck in Eq.~\ref{eq:router}. The router compresses the entire sequence into a single representation, losing positional information. Token-swap perturbations shift routing by only L2 = 0.22 (vs.\ L2 = 0.78 for token replacement), confirming the router is largely permutation-invariant. Per-token routing is the natural fix, but in our 26M-scale ablation it produced PPL 9.46 vs.\ 7.95 for mean-pooled routing ($+19\%$ worse), suggesting finer-grained routing introduces optimization difficulties at small scale. Whether this holds at 400M+ is unknown.

Within-head direction angles average 75.9\textdegree{} (vs.\ 89.4\textdegree{} for random vectors in 128D), indicating mild superposition \citep{elhage2022superposition}. The 576 directions span 88\% of the available capacity (effective rank 112.9/128, 83\% of pairs within 20\textdegree{} of orthogonal). The model packs directions tighter than random but avoids aggressive superposition, maintaining enough separation to prevent interference.

At 26M parameters, routing achieves 39.8\% PPL reduction on TinyStories and 18.1\% on FineWeb; at 400M, the reduction is 31--56\% on FineWeb. Parameter overhead is 3.9\% at both scales. Two data points are not a scaling law \citep{kaplan2020scaling, hoffmann2022training}, and we report them as preliminary evidence only.

\paragraph{Limitations.}
All results are from a single training run per model; variance across random seeds is unmeasured. We have tested at two scales only (26M and 400M); the method needs validation at 1B+. Our baseline is weaker than Pythia-410M due to $120\times$ less training data, so all efficiency claims are relative. A data-matched comparison (same tokens, 3.9\% more parameters in FFN instead of routing) is needed to isolate routing's contribution from inductive bias under data scarcity. Benchmark accuracy shows no improvement. The mean-pooling bottleneck restricts routing to sequence-level decisions. Circuit analysis covers factual recall and induction only; we have not tested syntax, coreference, arithmetic, or other circuit types. The Pythia perplexity comparison uses different tokenizers and is not directly interpretable. We provide no ablation on $K$ or router depth, the primary architectural hyperparameters. The domain classification result ($n{=}24$) is too small for statistical claims.

\section{Conclusion}

Directional routing adds dynamic feature suppression to transformer attention at 3.9\% parameter cost. On the two circuits we analyze, routing is the single non-redundant component while individual heads are interchangeable. The coordination mechanism is what matters. The architecture self-organizes into domain-adaptive early layers and syntactic pruning in late layers without any explicit pressure, and the 576 learned directions provide built-in interpretable features that are vocabulary-projectable and causally manipulable, complementary to sparse autoencoders.

The paper has clear gaps: no data-matched efficiency comparison, no ablation on $K$ or router depth, circuit analysis limited to two circuit types, benchmark accuracy flat, and a single training seed. Future work should address the mean-pooling bottleneck, validate at 1B+ parameters, run controlled efficiency experiments, and test routing's role in a broader set of circuits.

\bibliography{references}

@inproceedings{shazeer2017outrageously,
  title={Outrageously Large Neural Networks: The Sparsely-Gated Mixture-of-Experts Layer},
  author={Shazeer, Noam and Mirhoseini, Azalia and Maziarz, Krzysztof and Davis, Andy and Le, Quoc and Hinton, Geoffrey and Dean, Jeff},
  booktitle={International Conference on Learning Representations},
  year={2017}
}

@article{fedus2022switch,
  title={Switch Transformers: Scaling to Trillion Parameter Models with Simple and Efficient Sparsity},
  author={Fedus, William and Zoph, Barret and Shazeer, Noam},
  journal={Journal of Machine Learning Research},
  volume={23},
  number={120},
  pages={1--39},
  year={2022}
}

@inproceedings{lepikhin2021gshard,
  title={{GShard}: Scaling Giant Models with Conditional Computation and Automatic Sharding},
  author={Lepikhin, Dmitry and Lee, HyoukJoong and Xu, Yuanzhong and Chen, Dehao and Firat, Orhan and Huang, Yanping and Krikun, Maxim and Shazeer, Noam and Chen, Zhifeng},
  booktitle={International Conference on Learning Representations},
  year={2021}
}

@article{jiang2024mixtral,
  title={Mixtral of Experts},
  author={Jiang, Albert Q and Sablayrolles, Alexandre and Roux, Antoine and Mensch, Arthur and Savary, Blanche and Bamford, Chris and Chaplot, Devendra Singh and de las Casas, Diego and Hanna, Emma Bou and Bressand, Florian and others},
  journal={arXiv preprint arXiv:2401.04088},
  year={2024}
}

@article{dai2024deepseekmoe,
  title={{DeepSeekMoE}: Towards Ultimate Expert Specialization in Mixture-of-Experts Language Models},
  author={Dai, Damai and Deng, Chengqi and Zhao, Chenggang and Xu, R.X. and Gao, Huazuo and Chen, Deli and Li, Jiashi and Zeng, Wenge and Yu, Xingkai and Wu, Y. and others},
  journal={arXiv preprint arXiv:2401.06066},
  year={2024}
}

@inproceedings{cunningham2024sparse,
  title={Sparse Autoencoders Find Highly Interpretable Features in Language Models},
  author={Cunningham, Hoagy and Ewart, Aidan and Riggs, Logan and Huben, Robert and Sharkey, Lee},
  booktitle={International Conference on Learning Representations},
  year={2024}
}

@article{bricken2023monosemanticity,
  title={Towards Monosemanticity: Decomposing Language Models With Dictionary Learning},
  author={Bricken, Trenton and Templeton, Adly and Batson, Joshua and Chen, Brian and Jermyn, Adam and Conerly, Tom and Turner, Nick and Anil, Cem and Denison, Carson and Askell, Amanda and others},
  journal={Transformer Circuits Thread},
  year={2023}
}

@article{elhage2022superposition,
  title={Toy Models of Superposition},
  author={Elhage, Nelson and Hume, Tristan and Olsson, Catherine and Schiefer, Nicholas and Henighan, Tom and Kravec, Shauna and Hatfield-Dodds, Zac and Lasenby, Robert and Drain, Dawn and Chen, Carol and others},
  journal={Transformer Circuits Thread},
  year={2022}
}

@article{olsson2022context,
  title={In-context Learning and Induction Heads},
  author={Olsson, Catherine and Elhage, Nelson and Nanda, Neel and Joseph, Nicholas and DasSarma, Nova and Henighan, Tom and Mann, Ben and Askell, Amanda and Bai, Yuntao and Chen, Anna and others},
  journal={Transformer Circuits Thread},
  year={2022}
}

@inproceedings{wang2023interpretability,
  title={Interpretability in the Wild: a Circuit for Indirect Object Identification in {GPT}-2 Small},
  author={Wang, Kevin and Variengien, Alexandre and Conmy, Arthur and Shlegeris, Buck and Steinhardt, Jacob},
  booktitle={International Conference on Learning Representations},
  year={2023}
}

@inproceedings{meng2022locating,
  title={Locating and Editing Factual Associations in {GPT}},
  author={Meng, Kevin and Bau, David and Andonian, Alex and Belinkov, Yonatan},
  booktitle={Advances in Neural Information Processing Systems},
  year={2022}
}

@inproceedings{geva2023dissecting,
  title={Dissecting Recall of Factual Associations in Auto-Regressive Language Models},
  author={Geva, Mor and Bastings, Jasmijn and Filippova, Katja and Globerson, Amir},
  booktitle={Proceedings of the 2023 Conference on Empirical Methods in Natural Language Processing},
  year={2023}
}

@article{belinkov2022probing,
  title={Probing Classifiers: Promises, Shortcomings, and Advances},
  author={Belinkov, Yonatan},
  journal={Computational Linguistics},
  volume={48},
  number={1},
  pages={207--219},
  year={2022}
}

@inproceedings{kornblith2019similarity,
  title={Similarity of Neural Network Representations Revisited},
  author={Kornblith, Simon and Norouzi, Mohammad and Lee, Honglak and Hinton, Geoffrey},
  booktitle={International Conference on Machine Learning},
  year={2019}
}

@inproceedings{voita2019analyzing,
  title={Analyzing Multi-Head Self-Attention: Specialized Heads Do the Heavy Lifting, the Rest Can Be Pruned},
  author={Voita, Elena and Talbot, David and Moiseev, Fedor and Sennrich, Rico and Titov, Ivan},
  booktitle={Proceedings of the 57th Annual Meeting of the Association for Computational Linguistics},
  pages={5797--5808},
  year={2019}
}

@inproceedings{michel2019sixteen,
  title={Are Sixteen Heads Really Better than One?},
  author={Michel, Paul and Levy, Omer and Neubig, Graham},
  booktitle={Advances in Neural Information Processing Systems},
  year={2019}
}

@inproceedings{clark2019what,
  title={What Does {BERT} Look At? An Analysis of {BERT}'s Attention},
  author={Clark, Kevin and Khandelwal, Urvashi and Levy, Omer and Manning, Christopher D.},
  booktitle={Proceedings of the 2019 ACL Workshop BlackboxNLP},
  year={2019}
}

@article{kaplan2020scaling,
  title={Scaling Laws for Neural Language Models},
  author={Kaplan, Jared and McCandlish, Sam and Henighan, Tom and Brown, Tom B and Chess, Benjamin and Child, Rewon and Gray, Scott and Radford, Alec and Wu, Jeffrey and Amodei, Dario},
  journal={arXiv preprint arXiv:2001.08361},
  year={2020}
}

@inproceedings{hoffmann2022training,
  title={Training Compute-Optimal Large Language Models},
  author={Hoffmann, Jordan and Borgeaud, Sebastian and Mensch, Arthur and Buchatskaya, Elena and Cai, Trevor and Rutherford, Eliza and de Las Casas, Diego and Hendricks, Lisa Anne and Welbl, Johannes and Clark, Aidan and others},
  booktitle={Advances in Neural Information Processing Systems},
  year={2022}
}

@article{biderman2023pythia,
  title={Pythia: A Suite for Analyzing Large Language Models Across Training and Scaling},
  author={Biderman, Stella and Schoelkopf, Hailey and Anthony, Quentin and Bradley, Herbie and O'Brien, Kyle and Hallahan, Eric and Khan, Mohammad Aflah and Purohit, Shivanshu and Prashanth, USVSN Sai and Raff, Edward and others},
  journal={International Conference on Machine Learning},
  year={2023}
}

@article{penedo2024fineweb,
  title={The {FineWeb} Datasets: Decanting the Web for the Finest Text Data at Scale},
  author={Penedo, Guilherme and Kydl{\'i}{\v{c}}ek, Hynek and Lozhkov, Anton and Mitchell, Margaret and Raffel, Colin and Von Werra, Leandro and Wolf, Thomas},
  journal={arXiv preprint arXiv:2406.17557},
  year={2024}
}

@article{radford2019language,
  title={Language Models are Unsupervised Multitask Learners},
  author={Radford, Alec and Wu, Jeffrey and Child, Rewon and Luan, David and Amodei, Dario and Sutskever, Ilya},
  journal={OpenAI blog},
  year={2019}
}

@inproceedings{vaswani2017attention,
  title={Attention is All You Need},
  author={Vaswani, Ashish and Shazeer, Noam and Parmar, Niki and Uszkoreit, Jakob and Jones, Llion and Gomez, Aidan N and Kaiser, {\L}ukasz and Polosukhin, Illia},
  booktitle={Advances in Neural Information Processing Systems},
  year={2017}
}

@article{zellers2019hellaswag,
  title={{HellaSwag}: Can a Machine Really Finish Your Sentence?},
  author={Zellers, Rowan and Holtzman, Ari and Bisk, Yonatan and Farhadi, Ali and Choi, Yejin},
  journal={Proceedings of the 57th Annual Meeting of the Association for Computational Linguistics},
  year={2019}
}

@article{clark2018arc,
  title={Think you have Solved Question Answering? Try {ARC}, the {AI2} Reasoning Challenge},
  author={Clark, Peter and Cowhey, Isaac and Etzioni, Oren and Khot, Tushar and Sabharwal, Ashish and Schoenick, Carissa and Tafjord, Oyvind},
  journal={arXiv preprint arXiv:1803.05457},
  year={2018}
}

@inproceedings{bisk2020piqa,
  title={{PIQA}: Reasoning about Physical Intuition in Natural Language},
  author={Bisk, Yonatan and Zellers, Rowan and Le Bras, Ronan and Gao, Jianfeng and Choi, Yejin},
  booktitle={AAAI Conference on Artificial Intelligence},
  year={2020}
}

@inproceedings{sakaguchi2020winogrande,
  title={{WinoGrande}: An Adversarial {Winograd} Schema Challenge at Scale},
  author={Sakaguchi, Keisuke and Le Bras, Ronan and Bhagavatula, Chandra and Choi, Yejin},
  booktitle={AAAI Conference on Artificial Intelligence},
  year={2020}
}

@inproceedings{clark2019boolq,
  title={{BoolQ}: Exploring the Surprising Difficulty of Natural Yes/No Questions},
  author={Clark, Christopher and Lee, Kenton and Chang, Ming-Wei and Kwiatkowski, Tom and Collins, Michael and Toutanova, Kristina},
  booktitle={Proceedings of NAACL-HLT},
  year={2019}
}

@inproceedings{paperno2016lambada,
  title={The {LAMBADA} dataset: Word prediction requiring a broad discourse context},
  author={Paperno, Denis and Kruszewski, Germ{\'a}n and Lazaridou, Angeliki and Pham, Ngoc Quan and Bernardi, Raffaella and Pezzelle, Sandro and Baroni, Marco and Boleda, Gemma and Fern{\'a}ndez, Raquel},
  booktitle={Proceedings of the 54th Annual Meeting of the Association for Computational Linguistics},
  pages={1525--1534},
  year={2016}
}

@article{gao2021framework,
  title={A Framework for Few-Shot Language Model Evaluation},
  author={Gao, Leo and Tow, Jonathan and Abbasi, Baber and Biderman, Stella and Black, Sid and DiPofi, Anthony and Foster, Charles and Golding, Laurence and Hsu, Jeffrey and Le Noac'h, Alain and others},
  journal={Zenodo},
  year={2021},
  note={\url{https://github.com/EleutherAI/lm-evaluation-harness}}
}

@article{loshchilov2019decoupled,
  title={Decoupled Weight Decay Regularization},
  author={Loshchilov, Ilya and Hutter, Frank},
  journal={International Conference on Learning Representations},
  year={2019}
}
\bibliographystyle{plainnat}

\appendix

\section{Benchmark Results and Confidence Intervals}
\label{app:ci}

\begin{table}[H]
\centering
\caption{Multiple-choice benchmark accuracy \citep{gao2021framework}, $n{=}500$ examples each. BoolQ is the only benchmark where the 95\% CI on the difference excludes zero.}
\begin{tabular}{lcccc}
\toprule
Benchmark & Baseline & Routed & $\Delta$ & 95\% CI on $\Delta$ \\
\midrule
HellaSwag \citep{zellers2019hellaswag} & 32.2\% & 31.4\% & $-0.8$\% & $\pm 5.8\%$ \\
ARC-Easy \citep{clark2018arc} & 33.6\% & 32.0\% & $-1.6$\% & $\pm 5.9\%$ \\
ARC-Challenge \citep{clark2018arc} & 20.4\% & 24.7\% & $+4.3$\% & $\pm 5.2\%$ \\
PIQA \citep{bisk2020piqa} & 65.2\% & 64.2\% & $-1.0$\% & $\pm 5.9\%$ \\
WinoGrande \citep{sakaguchi2020winogrande} & 53.8\% & 51.8\% & $-2.0$\% & $\pm 6.2\%$ \\
BoolQ \citep{clark2019boolq} & 60.0\% & 53.0\% & $-7.0$\% & $\pm 6.1\%$ \\
LAMBADA \citep{paperno2016lambada} & 29.2\% & 28.6\% & $-0.6$\% & $\pm 5.6\%$ \\
\midrule
Average & 42.1\% & 40.8\% & $-1.2$\% & --- \\
\bottomrule
\end{tabular}
\end{table}

For $n{=}500$ independent trials with observed accuracy $p$, the 95\% CI on the difference between two independent proportions is $\Delta \pm 1.96 \sqrt{p_1(1{-}p_1)/n + p_2(1{-}p_2)/n}$.

Pythia-410M \citep{biderman2023pythia} results on the subset of benchmarks where we have comparable numbers: ARC-Challenge 21.3\% (ours: 24.7\% routed, 20.4\% baseline), PIQA 66.8\% (ours: 64.2\%, 65.2\%), WinoGrande 53.7\% (ours: 51.8\%, 53.8\%). Different training data, tokenizer, and evaluation harness limit direct comparison.

\section{Architecture and Efficiency Details}
\label{app:architecture}

\begin{table}[H]
\centering
\caption{Parameter breakdown for the routed model (433M total).}
\begin{tabular}{lrc}
\toprule
Component & Parameters & Share \\
\midrule
Feed-forward MLPs & 226,492,416 & 52.3\% \\
Attention (QKV + output proj) & 113,246,208 & 26.1\% \\
Embeddings (token + position) & 77,194,752 & 17.8\% \\
Router MLPs ($12 \times$ 4-layer) & 16,079,424 & 3.7\% \\
Direction vectors ($576 \times 128$) & 73,728 & $<$0.1\% \\
LayerNorm & 38,400 & $<$0.1\% \\
\midrule
Total & 433,124,928 & 100\% \\
\bottomrule
\end{tabular}
\end{table}

The top per-head routing contributions to the Paris logit are: L8H2 ($+4.14$, 103\% of the raw head magnitude), L9H4 ($+3.17$, 108\%), and L10H8 ($+2.55$, 107\%). In all three cases, the routing delta exceeds the raw head's own logit contribution, meaning routing is the dominant factor in each head's effect on the target token.

\section{Per-Head Routing Statistics}
\label{app:head-stats}

Individual heads in early layers show large cross-domain routing swings. L0H6 suppresses code at $w{=}0.59$ but factual at $w{=}0.35$ (a 0.24 swing). L0H4 suppresses code at 0.63 vs.\ prose at 0.44 (0.19 swing). In Layer~3, L3H2 shows a 0.14 swing between prose and math. These per-head differences drive the aggregate specialist counts reported in \S\ref{sec:two-regime}.

\section{Supplementary Figures}
\label{app:figures}

\begin{figure}[H]
    \centering
    \includegraphics[width=0.85\textwidth]{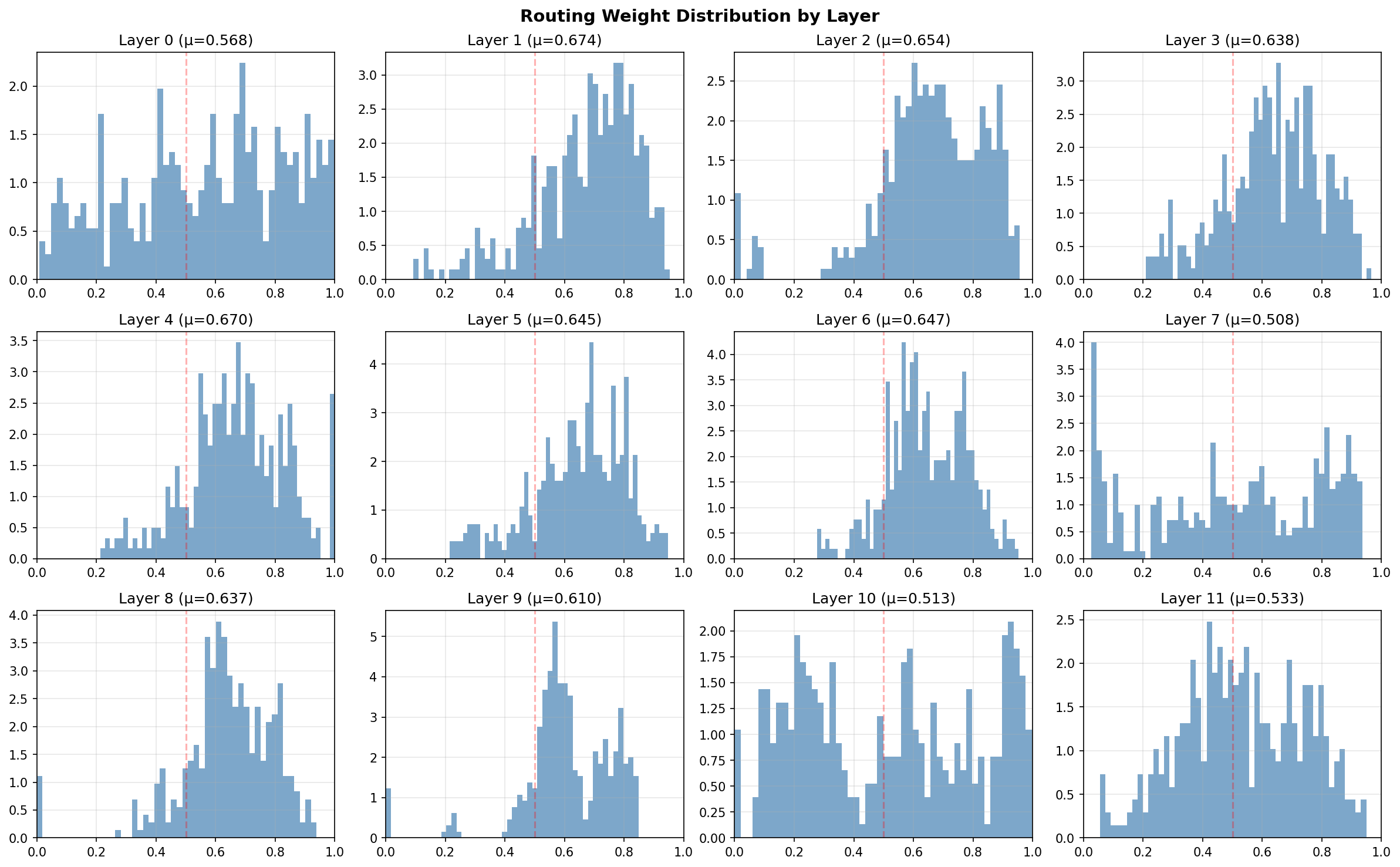}
    \caption{Per-layer routing weight distributions. Layer~0 shows the widest spread ($\sigma{=}0.272$), consistent with domain-adaptive specialist routing. Layer~7 is bimodal ($\sigma{=}0.290$, 14.8\% near-zero). Layer~9 never produces weights above 0.9 despite being the most critical layer. All layer means exceed 0.5 (range 0.508--0.674); the model prefers suppression over preservation on average.}
    \label{fig:routing-dist}
\end{figure}

\begin{figure}[H]
    \centering
    \includegraphics[width=0.85\textwidth]{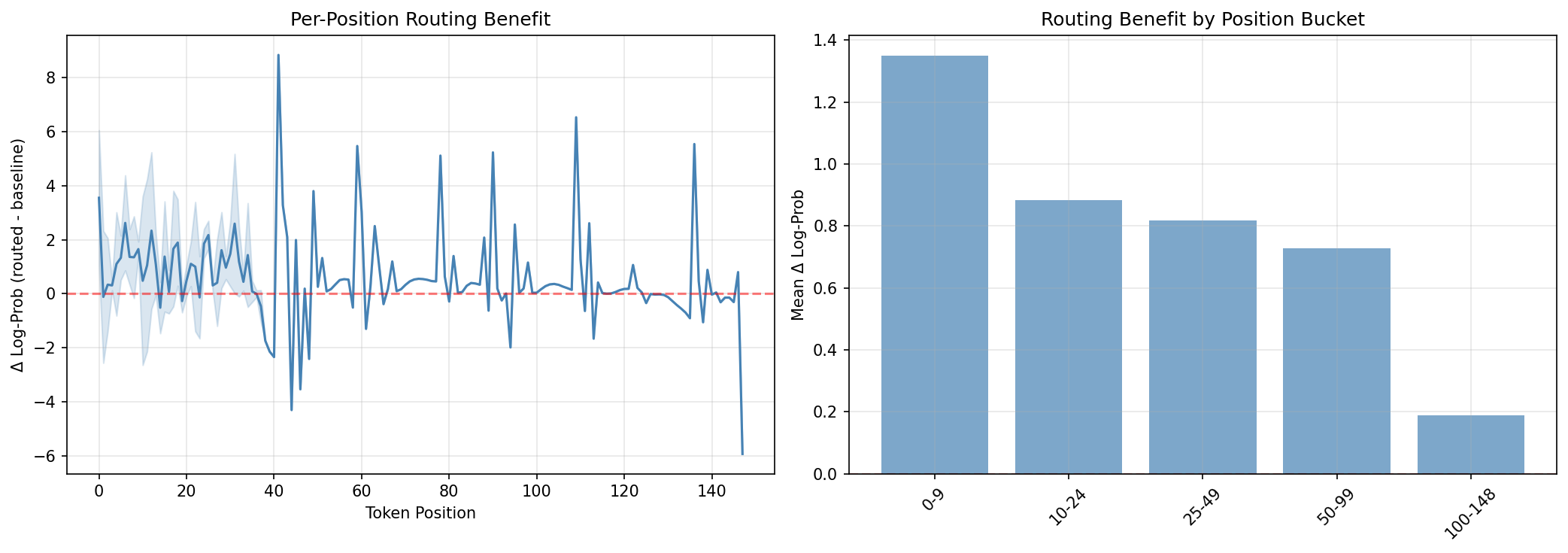}
    \caption{Routing benefit ($\Delta$ log-prob) by token position. Decays from $+1.35$ at positions 0--9 to $+0.19$ at 100+, a direct consequence of the mean-pooling bottleneck.}
    \label{fig:position-benefit}
\end{figure}

\begin{figure}[H]
    \centering
    \begin{subfigure}[t]{0.48\textwidth}
        \includegraphics[width=\textwidth]{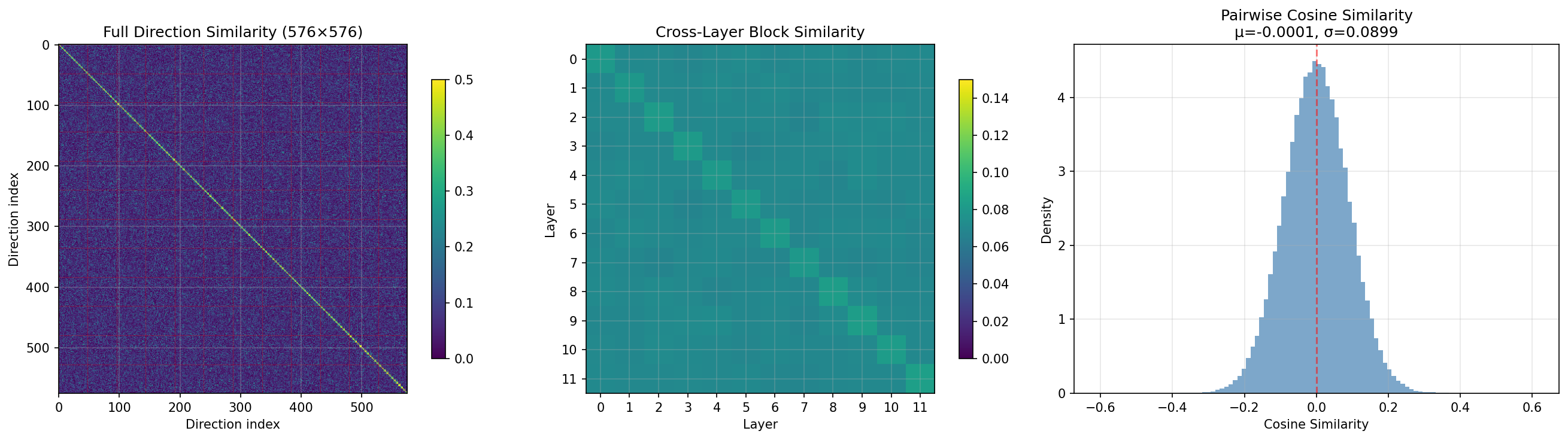}
        \caption{Direction cosine similarity matrix.}
    \end{subfigure}
    \hfill
    \begin{subfigure}[t]{0.48\textwidth}
        \includegraphics[width=\textwidth]{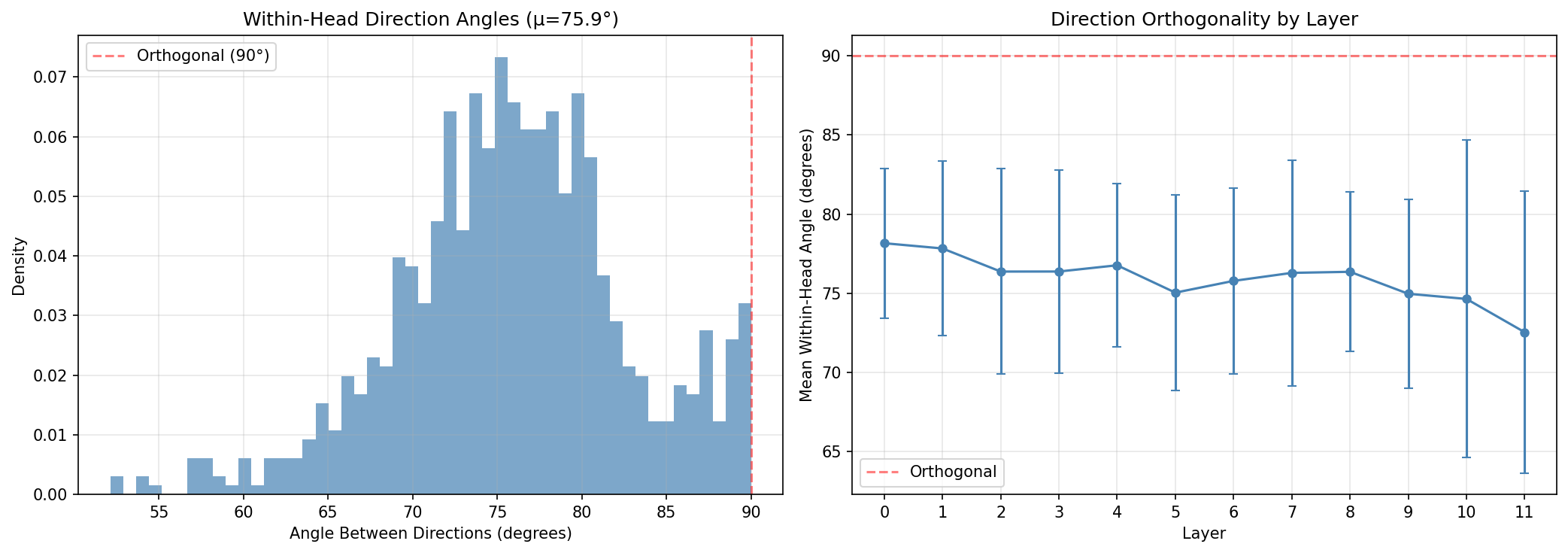}
        \caption{Within-head angle distribution vs.\ random baseline.}
    \end{subfigure}
    \caption{Direction geometry. Mean within-head angle: 75.9\textdegree{} (vs.\ 89.4\textdegree{} random in 128D). Effective rank: 112.9/128 (88\% capacity usage). Mild superposition, no dimensional collapse.}
    \label{fig:superposition}
\end{figure}

\begin{figure}[H]
    \centering
    \includegraphics[width=0.85\textwidth]{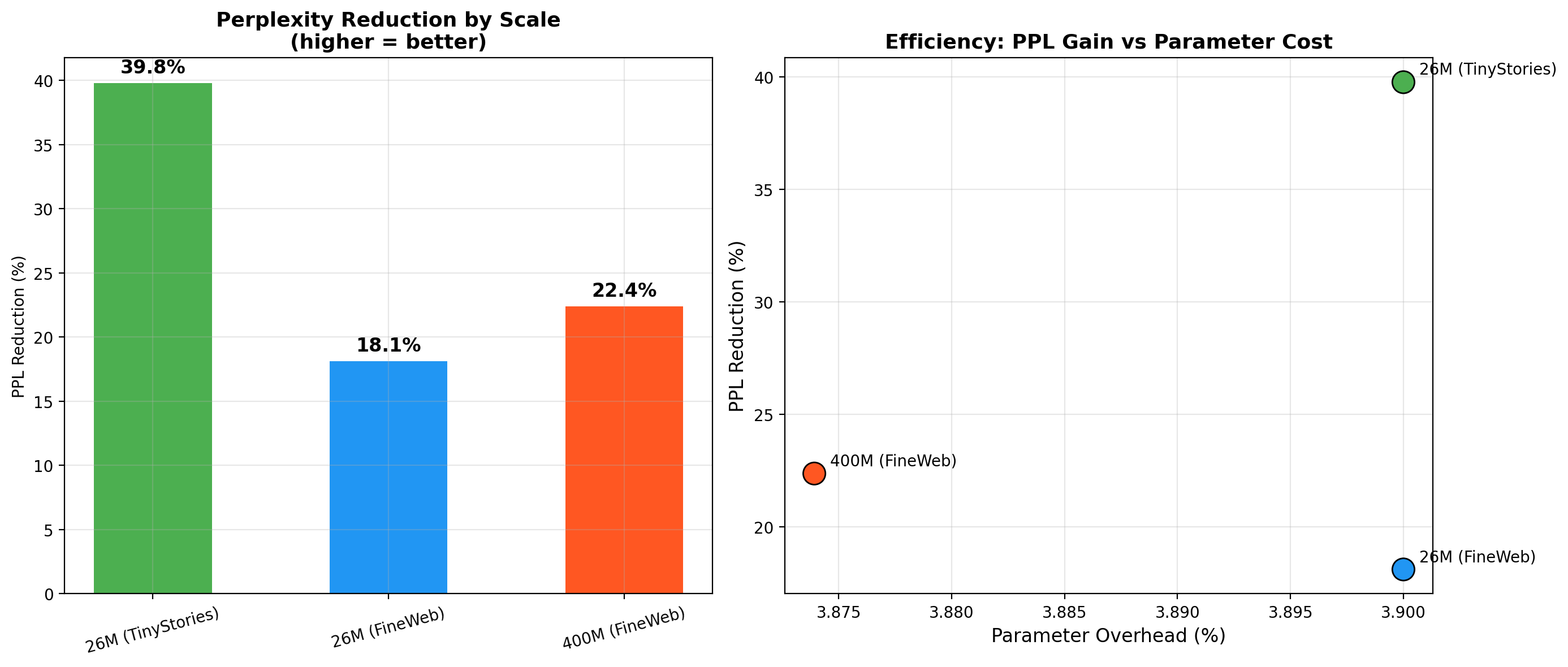}
    \caption{Preliminary cross-scale results at 26M and 400M parameters. Parameter overhead is 3.9\% at both scales. Two data points; not a scaling law.}
    \label{fig:cross-scale}
\end{figure}

\end{document}